# A Noise-robust Multi-head Attention Mechanism for Formation Resistivity Prediction: Frequency Aware LSTM

Yongan Zhang, *Student Member, IEEE*, Junfeng Zhao, Jian Li, Xuanran Wang, Youzhuang Sun, Yuntian Chen, *Member, IEEE* and Dongxiao Zhang, *Member, IEEE*

*Abstract*—The prediction of formation resistivity plays a crucial role in the evaluation of oil and gas reservoirs, identification and assessment of geothermal energy resources, groundwater detection and monitoring, and carbon capture and storage. However, traditional well logging techniques fail to measure accurate resistivity in cased boreholes, and the transient electromagnetic method for cased borehole resistivity logging encounters challenges of high-frequency disaster (the problem of inadequate learning by neural networks in high-frequency features) and noise interference, badly affecting accuracy. To address these challenges, frequency-aware framework and temporal anti-noise block are proposed to build frequency aware LSTM (FAL). The frequency-aware framework implements a dual-stream structure through wavelet transformation, allowing the neural network to simultaneously handle high-frequency and low-frequency flows of time-series data, thus avoiding high-frequency disaster. The temporal anti-noise block integrates multiple attention mechanisms and soft-threshold attention mechanisms, enabling the model to better distinguish noise from redundant features. Ablation experiments demonstrate that the frequency-aware framework and temporal anti-noise block contribute significantly to performance improvement. FAL achieves a 24.3% improvement in R2 over LSTM, reaching the highest value of 0.91 among all models. In robustness experiments, the impact of noise on FAL is approximately 1/8 of the baseline, confirming the noise resistance of FAL. The proposed FAL effectively reduces noise interference in predicting formation resistivity from cased transient electromagnetic well logging curves, better learns high-frequency features, and thereby enhances the prediction accuracy and noise resistance of the neural network model.

*Index Terms*—Formation resistivity, Frequency aware framework, noise-resistant block, well logs.

## I. INTRODUCTION

THE formation resistivity is an indispensable element for assessing and estimating underground physical field information. Therefore, the prediction of subsurface resistivity holds significant importance in various domains, such as the evaluation of hydrocarbon reservoirs, identification and assessment of geothermal energy resources, detection and monitoring of underground water bodies, as well as evaluation of the effectiveness of carbon capture and storage. In the evaluation of oil and gas reservoirs in petroleum exploration and development, the formation resistivity is used to assess the lithology, porosity, oil saturation, and other physical parameters, thereby evaluating the reservoir's storage capacity and the distribution of oil, gas [1, 2]. In groundwater detection and monitoring, the dynamic changes in formation resistivity can be employed to detect and monitor variations in underground water resources, enabling the assessment of the quantity and quality of groundwater resources [3, 4]. In the identification and evaluation of geothermal energy resources, the anomalies in formation resistivity are utilized to detect the presence and distribution of geothermal energy, minerals, and other geological resources [5, 6]. Regarding seismic prediction, the trend changes in formation resistivity can be analyzed to understand variations in the stress field of surrounding structures, facilitating the prediction of seismic events and their potential hazards [7, 8]. In the Carbon Capture, Utilization and Storage (CCUS), the sensitivity of formation resistivity is harnessed to monitor the migration and permeation of carbon dioxide during the CCUS process, assessing the effectiveness and safety of CCUS [9, 10]. Formation resistivity is primarily influenced by the type of liquid it contains, with conventional formations containing conductive saline water exhibiting significant differences in resistivity compared to formations filled with hydrocarbons. Therefore, accurate measurement of

This work was supported by the National Natural Science Foundation of China under Grant 62106116, Natural Science Foundation of Ningbo of China under Grant 2023J027 and the High Performance Computing Centers at Eastern Institute of Technology, Ningbo, and Ningbo Institute of Digital Twin.

Y. Zhang is with Ningbo Institute of Digital Twin and Eastern Institute of Technology, Zhejiang 315000, China, (Email: s21070021@s.upc.edu.cn)

J. Zhao is with Sinopec Matrix Corporation, Qingdao 266001, China, (Email: zhaojf.osjw@sinopec.com).

J. Li is with is with Ningbo Institute of Digital Twin and Eastern Institute of Technology, Zhejiang 315000, China, (Email: jli@idt.eitech.edu.cn).

X. Wang is with Sinopec Matrix Corporation, Qingdao 266001, China, (Email: wangxr.osjw@sinopec.com).

Youzhuang Sun is with China University of Petroleum（East China）, Qingdao 266580, China, (Email: 905736579@qq.com).

Y. Chen and D. Zhang are with Ningbo Institute of Digital Twin and Eastern Institute of Technology, Zhejiang 315000, China, (Email: cyt_cn@126.com and dzhang@eitech.edu.cn).

Supplementary materials are submitted.

*Corresponding author: Yuntian Chen and Dongxiao Zhang*

formation resistivity is of irreplaceable engineering value for locating hydrocarbon-bearing formations. The transient electromagnetic method (TEM) for through-casing resistivity logging technology addresses the limitation of casing wells, which cannot obtain resistivity through traditional logging methods applicable to open-hole logging. However, TEM data faces challenges such as noise in the data and the high-frequency disaster. These issues result in insufficient prediction accuracy in logging interpretation of TEM.

*A. Transient electromagnetic method over cased borehole logging technology*

The transient electromagnetic logging technology through casing resistivity logging is a new through-casing logging technique that compensates for the inability to obtain resistivity through traditional logging techniques (which are suitable for bare-hole logging) in cased wells, enabling the measurement of formation resistivity in cased wells. However, due to the extremely low resistivity of the metal casing compared to the formation resistivity (with the formation resistivity ranging from 1 Ω·m to 1000 Ω·m, while the typical value for metal casing resistivity is $2\times10^{-7}$ Ω·m), two challenges arise: (1) Amplification of environmental noise: Issues such as casing corrosion, blowouts, and current leakage can cause current fluctuations, and the low resistivity of the metal casing amplifies such interference noise; (2) Reduction of formation information: When there is a change in fluid within the formation, logging signals should detect current fluctuations and other information. The low resistivity of the metal casing reduces the proportion of such important information, leading to the masking of critical formation information under the influence of random noise such as electromagnetic interference.

In addressing the challenge of obtaining formation resistivity from cased boreholes, there has been relevant research. In 1939, Yeten [11] proposed estimating leakage current by measuring the differential voltage on the casing, introducing the concept of cased borehole transient resistivity logging to calculate formation conductivity. However, due to incomplete theories and methodologies at the time, this technology did not undergo extensive development and widespread application. In 1989, Kaufman and Vail [12] independently introduced methods for cased borehole resistivity logging and instrument design, considering the effects of casing conductivity variations and electrode position changes on detection results. They each implemented compensation measures, with Vail designing calibration and zeroing procedures directly into the instrument to address these effects. In 1990, Kaufman [13], based on theoretical studies and feature analysis of potentials, electric fields, and the second derivatives of potentials in cased boreholes, proposed an approximate theoretical model and detection theory based on transmission line equations. This laid the theoretical foundation for cased borehole resistivity logging methods and instrument design by establishing the relationship between the second derivative of potential and formation conductivity. In 1994, Schenkel et al. [14] introduced the Transmission Line Equation (TLE), which calculated numerical solutions for potentials and their derivatives by incorporating finite-length casing, variable-radius annular zones, and vertical discontinuities to calculate formation resistivity, achieving an accuracy of up to 60%. In 2006, Pardo et al [15]. simulated measurement results under different frequencies using a multi-frequency cased borehole resistivity tool through a goal-oriented hp-adaptive finite-element method to evaluate rock properties. Experimental results demonstrated the capability of this method for logging simulations in the presence of highly conductive casings, but numerical calculations from empirical formulas proved inadequate in certain scenarios, leading to insufficient accuracy. In 2018, Liu et al. [16] addressing the limitation of the conventional TLE in calculating resistivity logging responses in fractured formations through casing, proposed a new method utilizing both TLE and the current flux tube model. This approach considered fracture conditions and extended to the calculation of lateral formation resistivity. However, it still falls within the realm of empirical formulas, resulting in limitations in accuracy and applicability. In 2023, Liu et al. [17] addressing the inability of the conventional TLE to investigate current and potential distributions in borehole fluids, proposed an improved TLE. The enhanced TLE incorporates formation information outside the casing and fluid inside the well, allowing for the analytical solution of casing wall potential distribution. Despite the improvement, the enhanced TLE is still limited to analytical solutions under theoretical conditions, making it unsuitable for all operational scenarios and subject to limitations in accuracy and applicability. Presently, cased borehole transient electromagnetic logging curves present low signal-to-noise ratios and high redundancy, making traditional empirical formula calculations and linear fittings unable to extract sufficient data features, leading to challenges in predicting formation resistivity from cased borehole logging curves. Therefore, addressing the inadequacies of traditional methods in predicting formation resistivity from cased borehole transient electromagnetic logging curves, this study proposes improvements using a modified time-series neural network model.

*B. Reservoir interpretation based on artificial intelligence in well logging*

Predicting formation resistivity from logging curves is a logging interpretation problem. For logging interpretation problems, artificial intelligence (AI) technologies have been successfully applied to various aspects of series data problem [18], [19]. Numerous studies have demonstrated the applicability of AI in handling diverse data sources in the field of well logging [20-23] and petrophysics [24-26].

In 2010, Karimpouli et al. [27] developed a supervised Committee Machine Neural Network (SCMNN) voting model for reconstructing petrophysical parameter curves. In 2016, Korjani et al. [28] introduced a reservoir modeling method based on deep neural networks to predict rock physics features. Virtual log data for well segments lacking log curves and core data were generated using abundant geological data from nearby wells. In 2017, Parapuram et al. [29] proposed a multi-stage curve generation scheme where each stage's log curve serves as a constraint in predicting the next stage's curve, enhancing the reliability of the final interpretation. In 2018, Akinnikawe et al. [30] proposed a practical approach involving variable clustering analysis and data transformation to differentiate input features in well log data. This process enables the training and testing of various machine learning methods. In 2018, Zhang et al. [31] utilized a cascaded long short-term memory (CLSTM) neural network for the reconstruction of logging curves, introducing multiple layer mapping relationships to enhance the accuracy of logging curve reconstruction. This study was the first to employ temporal neural networks to learn the temporal characteristics of logging curves. Wu et al. [32] introduced a workflow combining cross-entropy clustering, Gaussian Mixture Model, and Hidden Markov Model to propagate information from training wells to other wells by identifying locally stationary clustering clusters corresponding to formation zones. In 2018, Akkurt et al. [33] built an automated process for rapid learning and enhanced analysis of rock physics to detect outliers in well logging data, evaluate inter-well similarity post-outlier removal using clustering information, and subsequently reconstruct well log curves. In 2020, Chen et al. [34] proposed an ensemble Long Short-Term Memory (EnLSTM) network capable of processing sequential data on a small dataset. By combining ensemble neural network and cascaded LSTM network, EnLSTM reduces costs and saves time. In the same year, Chen et al. [35] introduced a physics-constrained LSTM, incorporating the physical mechanism behind geomechanically parameters as prior information. This model directly estimates geomechanically logs based on easily available data, demonstrating improved prediction accuracy by leveraging domain knowledge. In 2023, Sun et al. [36] developed the Inception-BiGRU model, a Gated Recurrent Unit Network based on the Inception Module, to improve the accuracy of missing log value prediction by extracting spatial relationships within well log curves.

While AI applications have propelled well logging curve predictions beyond challenges of multi-source data learning and deep feature extraction, limitations persist in dealing with low signal-to-noise ratios and high redundancy of well logging curves. Addressing the prediction of formation resistivity from cased borehole transient electromagnetic logging curves, this study proposes the establishment of a temporal anti-noise block to enhance the prediction accuracy and model generalization performance by learning from low signal-to-noise ratio, high redundancy multi-source well logging data. Therefore, the frequency-aware framework and temporal anti-noise Block are proposed to address the high-frequency disaster problem and noise issue in neural network learning.

*C. High-frequency disaster*

The prediction of formation resistivity from well logging curves falls within the domain of scientific machine learning problems [37], [38], arising challenges of frequency principle and high-frequency disaster. Frequency principle refers to the learning sequence principle in the frequency space of fitting training samples by neural networks, where learning progresses from low-frequency to high-frequency. This results in the neural network inadequately learning high-frequency features, known as the high-frequency disaster. In scientific machine learning problems, existing research has demonstrated the presence of high-frequency disaster in various applications such as solving partial differential equations (PDEs) [39], image generation [40], function fitting [41], among others, governed by frequency principles [42] and high-frequency disaster [43].

In 2015, LeCun et al. [44] observed that the general principles underlying the highly non-convex optimization problem in deep neural networks (DNNs) remained unclear. In 2019, Xu et al. [42] demonstrated a universal Frequency Principle in DNNs, revealing that these models often fit target functions from low to high frequencies. Experimental results indicated that most deep learning models exhibit frequency principles in learning temporal features, leading to the challenge of high-frequency disaster and ultimately reducing model prediction accuracy. In 2020, Liu et al. [43] addressed the high-frequency disaster problem caused by the frequency principle by proposing a Multi-scale Deep Neural Network. Experimental results demonstrated the superiority of this model over traditional fully connected DNNs, serving as an effective mesh-less numerical method for Poisson-Boltzmann equations with ample frequency contents over complex and singular domains. In 2023, Song et al. [45] addressed challenges in the training cost and low solution accuracy during the model equation encoding process in physics-informed neural networks. They introduced frequency domain learning and proposed a frequency domain physics-informed neural network. Experimental results demonstrated a significant reduction in solution errors by 1 to 2 orders of magnitude and a simultaneous improvement in training efficiency by 6 to 20 times. In the well logging curves, the temporal dependency of data involves both high-frequency and low-frequency components. The primary sources of signal features in the data are the longitudinal homogeneity of reservoir fluids, rock types, and formation pore structures. Therefore, through-casing well logging curve signals exhibit a characteristic where the signal strength attenuates with increasing frequency. This leads to the frequency principle in well logging curves, where in the computation of the gradient descent in neural network models, low-frequency signals contribute more significantly to the gradient than high-frequency signals. In this scenario, the fitting degree of high low-frequency signals can negatively impact the learning ability of well logging curves for high-frequency variations in formation characteristics. This difficulty arises from challenges in predicting the mapping relationships of high-frequency variations in fluid properties, rock types, and pore structure changes. This phenomenon results in the high-frequency disaster in well logging curve predictions.

In the resistivity logging curve prediction, a salient question is how to address high-frequency disaster and noise effects. Therefore, this paper proposes a frequency-aware framework and a temporal anti-noise Block to respectively solve the high-frequency disaster problem and noise issue produced by neural network learning. Overall, the structure of this paper is divided into three parts: in the Introduction section, the TEM method and the two challenges that exist are introduced; in the Methodologies section, the proposed framework and block as well as the Frequency aware LSTM are present; in the Experiments and Results section, the Missing well logs prediction experiment and the Robustness analysis with noisy data experiment are introduced.

## II. METHODOLOGIES

This section consists of three subsections introducing the two proposed modules and the assembled temporal neural network model. In Section 2.1, the specific structure of the frequency-aware framework and its application of wavelet transform and dual-stream architecture are presented to address challenges related to high-frequency disaster in well logging curve learning. Section 2.2 provides an in-depth exploration of the temporal anti-noise block, explaining its structure and the principles behind the application of soft-threshold processing units to mitigate the impact of noise and redundancy in transient electromagnetic well logging curves. Section 2.3 focuses on the specific structure of the Frequency aware LSTM (FAL), illustrating how proposed framework and block are combined.

### A. Frequency-aware framework

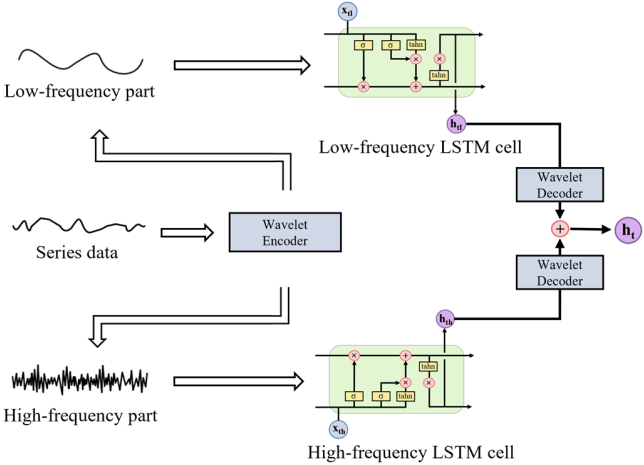

**Fig. 1.** Frequency-aware framework (FAF, the upper part is for low-frequency flow processing, and the lower part is for high-frequency flow processing).

To address the frequency principle characteristics and the challenge of high-frequency disaster in well logging curve data, frequency-aware framework (FAF, as illustrated in the figure below) based on wavelet transform is proposed. The FAF primarily consists of a wavelet encoder that utilizes wavelet transform to divide the temporal data into two streams: high-frequency flow features and low-frequency flow features. Both high-frequency and low-frequency flow features undergo separate temporal neural networks before being output to their respective wavelet decoders, also based on wavelet transform. After transformation into high-frequency and low-frequency flows, the results are combined to form the prediction outcome. The FAF segregates the high-frequency and low-frequency flows of temporal data. Through the dual-stream structure, it simultaneously provides feedback and adjusts both high-frequency and low-frequency flows, thereby circumventing the frequency principle, combating the challenge of high-frequency disaster, and ultimately improving prediction accuracy.

Wavelet Transform is a novel analytical method that inherits and develops the localized idea of short-time Fourier transform. Simultaneously overcoming drawbacks such as fixed window size in relation to frequency changes, WT provides an ideal tool for signal time-frequency analysis and processing by offering a "time-frequency" window that varies with frequency. By employing wavelet transform, the frequency domain features of through-casing well logging curves can be effectively highlighted, enabling localized analysis of both time and frequency characteristics. Through this approach, the input data is transformed from time-domain to frequency-domain by the wavelet encoder in advance, achieving frequency decomposition for multiscale refinement and dividing the input data into high-frequency and low-frequency parts. Subsequently, within the dual-stream results, the high-frequency and low-frequency parts are input into two different LSTM Cells as high-frequency flow and low-frequency flow, respectively, to learn high-frequency and low-frequency features. Finally, the wavelet decoder transforms the feature signals from frequency-domain data back to time-domain data and facilitates feedback regulation learning.

The FAF based on wavelet transform separates high-frequency and low-frequency features through wavelet transform. It simultaneously engages in gradient descent and feature learning for both high-frequency and low-frequency data, thereby enabling the model to, to a certain extent, counteract the frequency principle and avoid high-frequency disaster.

The encoding formula for the wavelet transform is expressed as follows:

$$Ws(a,b) = |a|^{-1/2} \int_R s(t) \bar{\Psi}\left(\frac{t-b}{a}\right) dt \quad (1)$$

$$\Psi_M(t) = e^{i\omega_0 t} e^{-\frac{t^2}{2}} \quad (2)$$

where $Ws(a,b)$ represents the result of continuous wavelet transform, s(t) is a variable representing the time t of the input series data. The parameter a is the scaling parameter, related to the frequency, and b is the position parameter. $\bar{\Psi}(t)$ is the complex conjugate function of $\Psi(t)$. $\Psi_M(t)$ is the wavelet function. In dealing with series data, a non-orthogonal Morlet wavelet is chosen. ω0 is a constant representing the center frequency, and i is the imaginary unit determining whether the information in the transformation result is in the time domain or spatial domain.

The Mallat algorithm [46] is employed for wavelet transform to decompose low-frequency data and high-frequency data. The Encoder and Decoder for wavelet transform are illustrated in the figure below:

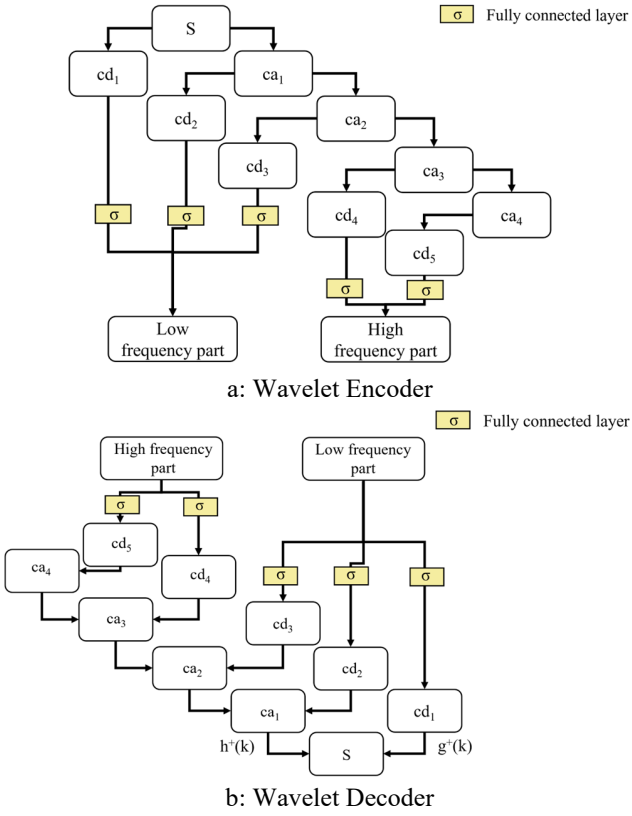

**Fig. 2.** Wavelet transforms (S, ca, cd represent sequence data, approximation coefficients, and detail coefficients)

where the calculation formulas for the approximation coefficient $a_{j-1,k}$ and the detail coefficient $d_{j-1,k}$ are expressed as follows:

$$\begin{cases} a_{j-1,k}(n) = \sum_{k=-\infty}^{+\infty} a_{j,k} h(k-2n), n=0,1,...,2^{N-j}-1 \\ d_{j-1,k}(n) = \sum_{k=-\infty}^{+\infty} a_{j,k} g(k-2n), n=0,1,...,M-1 \end{cases} \quad (3)$$

where h(k) represents the impulse response of the low-pass filter H, g(k) is the impulse response of the high-pass filter G, M denotes the total number of wavelet decomposition levels (set to 5 in the model), and j is the current wavelet decomposition level.

The reconstruction expression is given by:

$$\begin{cases} a_j = \sum_{k=-\infty}^{+\infty} \{h^+(n-2k)a_{j-1,k} + g^+(n-2k)d_{j-1,k}\} \\ j = M-1, M-2, ..., 0 \end{cases} \quad (4)$$

where $h^{+(k)}$ represents the impulse response of the reconstruction low-pass filter, and $g^{+(k)}$ is the impulse response of the reconstruction high-pass filter.

### B. Temporal anti-noise Block

In this study, there are 35 input curves in the transient electromagnetic logging curve of the casing. These curves exhibit highly redundant and low signal-to-noise ratio characteristics in the data. The high redundancy refers to the presence of a large amount of repeated or low-value information in the dataset, while the low signal-to-noise ratio indicates that the useful information in the dataset is relatively low compared to the noise. Therefore, the construction of the temporal anti-noise block is proposed as a neural network module tailored for addressing the challenges posed by low signal-to-noise ratio and high redundancy in the data.

Temporal anti-noise block serves as a denoising block, specifically designed for data with strong noise and excessive redundancy. Structurally, Temporal anti-noise block consists of three parts: residual unit, threshold learning unit, and soft-threshold processing unit. The residual unit (The green part of Figure 3) primarily reduces training difficulties through identity mapping. The threshold learning unit (The blue part of Figure 3), guided by attention mechanisms, learns appropriate thresholds through feedback adjustment and feeds it into the soft-threshold processing unit as the input of the soft threshold formula. The soft-threshold processing unit (The red part of Figure 3) is a crucial part for noise data removal, implementing the soft-threshold method [47] to set features within a specific range to zero, thereby eliminating specific features (either noise or redundant features). The soft thresholding method is a denoising method that can automatically learn and adapt to various denoising rules under different noise distributions based on the formula to adjust the threshold value. Essentially, Temporal anti-noise block combines residual networks, soft-thresholding functions, and attention mechanisms. It introduces soft-thresholding into the residual module as a shrinkage layer, employs multi-head attention mechanisms to set adaptive thresholds, and ensures that the model, through feedback adjustment, focuses on repeatedly occurring features in multidimensional redundant information. It forms a weight matrix submitted to the soft-threshold processing unit, reducing the impact of noise interference and redundant information on feature extraction in input samples.

The residual unit, soft-threshold processing unit, and deep residual shrinkage network share similarities with the deep residual shrinkage network [48]. The expression for the soft-threshold function and its derivative are provided in the following formulas

$$\text{soft}(x,T) = \begin{cases} x+\tau & x < -\tau \\ 0 & |x| \leq \tau \\ x-\tau & x > \tau \end{cases} \quad (5)$$

$$\frac{\partial y}{\partial x} = \begin{cases} 1 & x > -\tau \\ 0 & |x| \leq \tau \\ 1 & x > \tau \end{cases} \quad (6)$$

where $\tau$ represents the threshold size in the soft-threshold function, x is the input, and y is the output.

The threshold learning unit incorporates the idea of multi-head attention mechanism [49], and its specific structure is illustrated in the following figure:

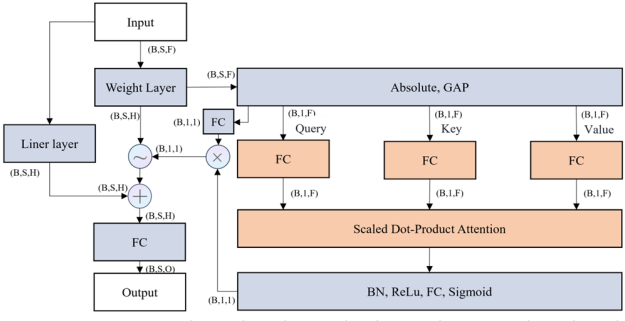

**Fig. 3.** Temporal anti-noise Block (red part: the threshold learning unit, blue part: the soft-threshold processing unit)

where B, S, and F denote Batch size, sequence length, and feature quantity, respectively. Similar to the soft-thresholding under the deep attention mechanism of the deep residual shrinkage network, temporal anti-noise block primarily employs soft-thresholding to set irrelevant features to zero in the feature vector, thereby filtering out features relevant to the current task to address the high redundancy in the data. Initially, the input multidimensional sequence data undergoes absolute value transformation and global mean pooling to extract global features. These processed features are then input into the multi-head attention mechanism for temporal feature extraction, learning a set of weights to suppress noise, which are subsequently fed into the soft-thresholding processing unit. In the soft-thresholding processing unit, the obtained weights and the temporal data undergo soft-thresholding calculations using formulas (5) and (6), suppressing irrelevant strong noisy features to counteract the high-noise characteristics of the data. Finally, the results of the soft-thresholding calculations are added to the residual results and fed into the fully connected layer. This process ensures that the entire neural network, during the learning of noise filtering, mitigates gradient vanishing and feature elimination. In summary, temporal anti-noise block has two main characteristics:

(1) Noise and feature filtering in time-series data: Addressing the issue of excessive redundant information and the interference caused by weak and noisy features in well logging input data, the block introduces a multi-head attention mechanism for noise filtering. The multi-head attention mechanism in the threshold learning unit performs feature selection and noise filtering, allowing the model to focus on removing noise and selecting relevant features after feedback adjustment.

(2) Multi-attention mechanism: Combating the problem of insufficient feature learning, the model combines both multi-head attention mechanism and soft-thresholding attention mechanism. The multi-head attention mechanism is integrated into the soft-thresholding recognition process and assists in setting adaptive thresholds. By identifying repeated features in multidimensional time-series data, it learns a weight matrix capable of removing noise and redundant information, which is then submitted to the soft-thresholding processing unit. Therefore, the model's dual attention mechanisms, coupled with residual and contraction operations, reinforce both the deep learning and sparse representation capabilities of the network.

### C. Frequency aware LSTM (FAL)

Addressing the challenges posed by low signal-to-noise ratio, high redundancy in through-casing electromagnetic well logging curves, and the frequency principle and high-frequency disaster issues associated with conventional neural network feedback learning, the FAF and Temporal anti-noise block are combined with the conventional temporal neural network LSTM. Firstly, the temporal anti-noise block and LSTM are combined to propose temporal anti-noise LSTM block (TAL), the structure is shown below:

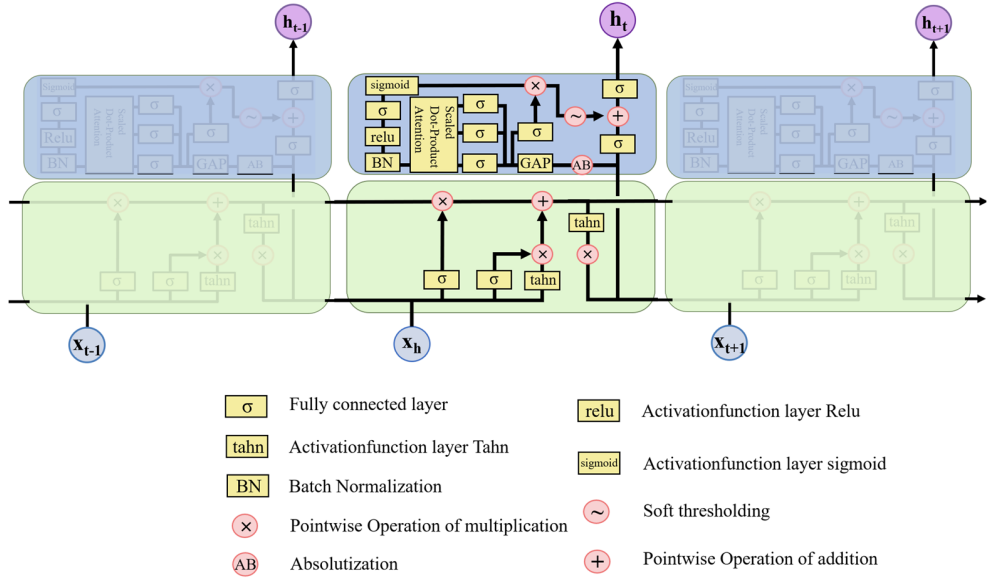

**Fig. 4.** Temporal anti-noise LSTM (TAL, blue part is temporal anti-noise block and green part is conventional LSTM cell)

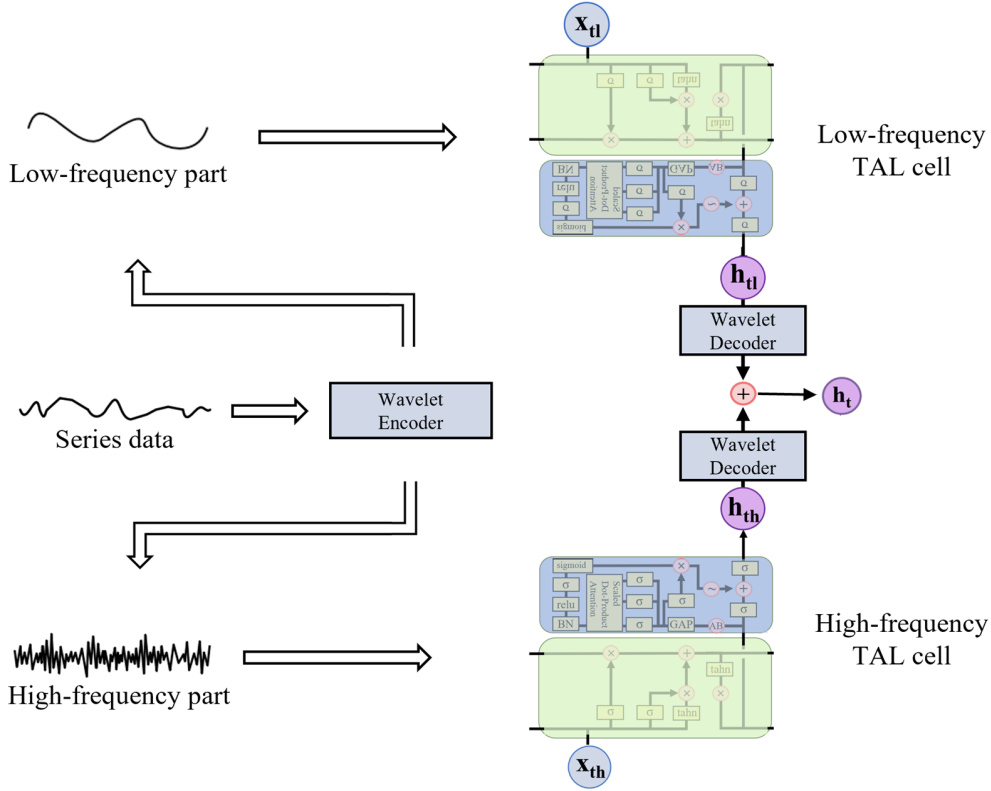

**Fig. 5.** Frequency aware LSTM (FAL, it is different from FAF that TAL cells replace LSTM cells)

In the Figure 4, input data is initially processed through a conventional LSTM cell (Figure 4 in green). Subsequently, it enters the temporal anti-noise block (Figure 4 in blue), where the multidimensional sequence data undergoes absolute value transformation and global average pooling to extract global features. This transformed data is then input into the multi-head attention mechanism for temporal feature extraction. The soft threshold calculation is performed using (5) and (6), and the resulting soft threshold calculation is added to the residual output before entering the fully connected layer. Finally, the loss is calculated with the label data, and adjustments are made through feedback. In comparison to LSTM, the additional temporal anti-noise block possesses a depth attention mechanism with multi-head attention and soft thresholding, handling the selection of multidimensional features and suppression of temporal data noise, respectively. The residual structure enables TAL to counteract gradient vanishing and feature elimination when filtering strong noise. Figure 4 illustrates the complete structure of TAL, and Figure 5 presents the structure of Frequency aware LSTM (FAL).

In FAL, analogous to the structure of FAF, temporal data is transformed from the time domain to the frequency domain, specifically into high-frequency and low-frequency components, through a wavelet transform encoder. The dual-stream architecture of FAL is employed to handle the high-frequency and low-frequency components, where the neural networks in the FAL dual-stream structure are based on TAL. The network is trained to learn features from the high-frequency data stream and low-frequency data stream separately. The output of the dual-stream structure undergoes reverse transformation using a wavelet transform decoder to convert back to the time domain. Subsequently, a loss function is computed by comparing the output with the label data of normal temporal sequences, and feedback adjustments are made. In comparison to TAL, the additional wavelet transforms encoder and decoder in FAL enable the extraction of high-frequency and low-frequency features. Through simultaneous learning from high-frequency and low-frequency features by the dual-stream structure that consisted with TAL, FAL not only avoids high-frequency disaster, but also reduces the impact of noise.

### III. EXPERIMENTS AND RESULTS

This section comprises data pretreatment and two experiments. In Section 3.1, the specific details of the data and the preprocessing pipeline will be elucidated, including dataset allocation, training methodology, parameter settings, etc. In Section 3.2, the experiment of missing well logs prediction. To compare the effectiveness of each module in the model, an ablation experiment against conventional models is conducted, analyzing the roles of individual modules. Furthermore, to showcase the model's capability in addressing high-frequency disaster, the performance results in both the high-frequency domain and low-frequency domain are presented, analyzing the effects of various modules on different frequency components of the data. In Section 3.3, the experiment of robustness analysis with noisy data is introduced. To evaluate the modules' ability to mitigate the impact of noise, the models are tested on datasets with two common types of noise added. By analyzing the decrease in predictive accuracy for each model, the capacity of the proposed modules in handling noise interference is examined.

*A. Data and Pretreatment*

The dataset is derived from the alluvial plains in eastern China. The input dataset comprises through-casing electromagnetic logging curve data, conventional well logging data, while the output consists of formation resistivity data obtained from barefoot well logging curves. The primary focus is on predicting formation resistivity based on through-casing electromagnetic logging curves. Due to the multi-point distributed measurement instruments of through-casing electromagnetic logging, simultaneous measurements are essential to ensure data accuracy. Therefore, the number of through-casing electromagnetic logging curves is relatively greater compared to conventional well logging curves. In this study, the dataset consists of three wells, encompassing a total of 35 through-casing electromagnetic logging curves and four conventional well logging curves. The prediction target is one set of formation resistivity data. The conventional well logging curves include AC (acoustic logging curve), GR (natural gamma curve), CON (induction logging curve), and SP (spontaneous potential curve). The through-casing electromagnetic logging curves include STEMUB01~STEMUB35 (readings from instrument current meters 01 to 35), as well as RT (formation resistivity).

To illustrate the low signal-to-noise ratio and high redundancy characteristics of the multi-source data, a correlation analysis of through-casing transient electromagnetic logging curve data was conducted. The following figure presents the correlation results between 42 through-casing transient electromagnetic logging curves and formation resistivity.

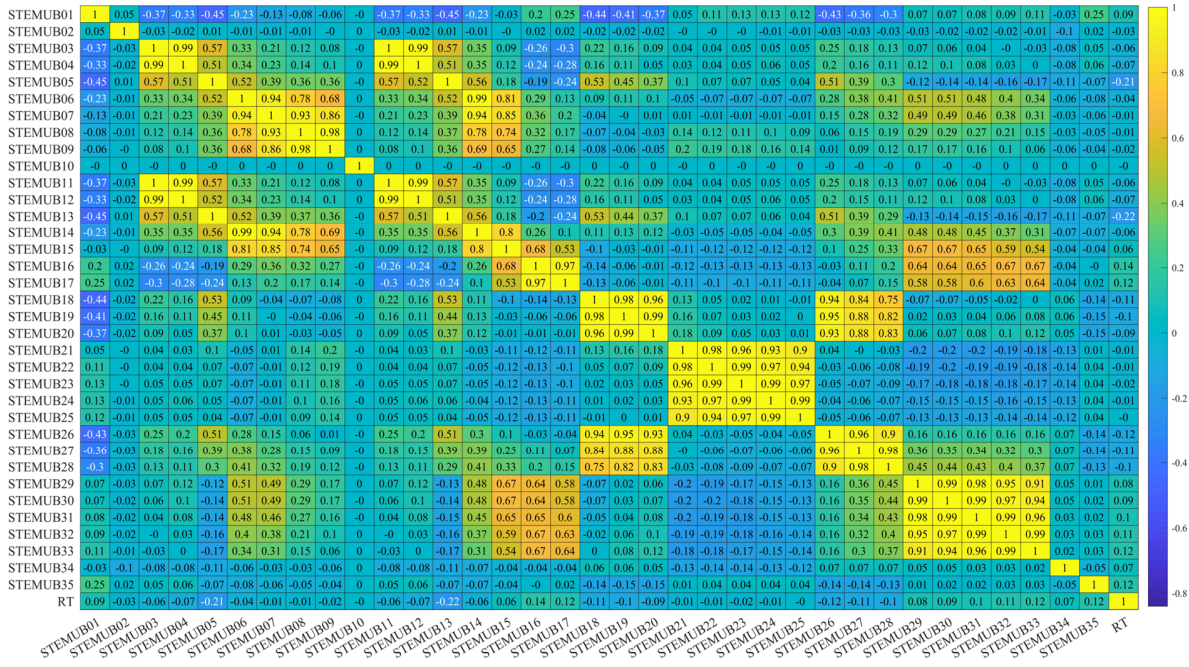

**Fig. 6.** Correlation between electromagnetic logging curve through casing and formation resistivity

The data preprocessing involves two main components: granularity alignment and standardization.

(1) Granularity Alignment: Primarily focusing on the granularity of through-casing transient electromagnetic logging curves, the conventional well logging curves (granularity of 0.01m) are sparsely aligned with and adjusted to match the granularity of the through-casing transient electromagnetic logging curves (granularity of 0.125m). The entire process ensures depth consistency, as detailed in follow:

$$s_i = t_{8j} \qquad (7)$$

where $s_i$ represents the value corresponding to the index $i$ of the sparsely processed data, where the granularity is 0.125m. $t_{8j}$ denotes the value corresponding to the index $8j$ of the original data, with a granularity of 0.01m.

(2) Standardization: Standardization is performed using conventional normalization methods, as specified in fellow:

$$X_{std} = \frac{X - u}{\sigma} \qquad (8)$$

where $X_{std}$ represents the standardized data, $X$ is the original data, $u$ is the mean of the data, and $\sigma$ is the standard deviation of the data.

*B. Experiment setting*

Before model training, we employed a time window method [50] with depth as the sequence dimension for data training and prediction, as illustrated in the figure below. The sequence length is determined by the reservoir homogeneity of the logging data source, and for this experiment, the sequence length is set to 5m (equivalent to the length of 40 samples). Figure 7 depicts the correspondence between the training data samples and labels in the experiment. The "Train Window" represents the sliding window in the time window method, sliding at intervals of one sample point each time. The data within each window serves as an independent sample in each training iteration, and the corresponding target data for the depth sequence acts as an independent label. Since the experimental sequence output results involve repetitive predictions at various depths, the mean of the output results at each depth in this experiment is considered as the true output result.

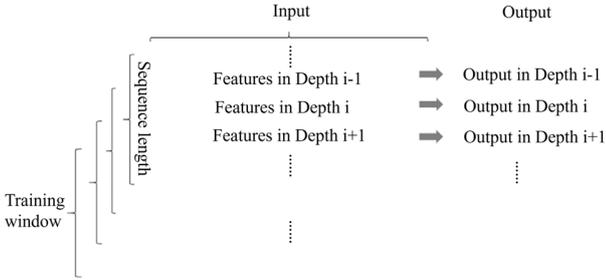

**Fig. 7.** Sequence prediction methods for depth sequences data

In terms of model hyperparameters, the sequence length is set to 40 (corresponding to a depth interval of 5m), the batch size is set to 64, and the number of features is 50 (indicating the number of input well log curves). The parameter settings are summarized in the following table.

TABLE I
HYPERPARAMETERS OF OUR MODELS IN EXPERIMENTS

| Parameter | Setting |
| --- | --- |
| Optimizer | Adam |
| Loss Function | Mean Absolute Error (MAE) |
| Learning Rate | {0.001} |
| Batch Size | 64 |
| Sequence Length | 40 |

The sequence length is set to 40, the batch size is configured as 64, and the number of features is specified as 50. The model training is terminated when the training loss stabilizes (change less than 0.001). The parameters of the network structure are outlined in the following table (BN: Batch Normalization layer, ReLU: Rectified Linear Unit layer, FC: Fully Connected layer, GAP: Global Average Pooling layer). The sequence length is set to 40, the batch size is configured as 64, and the number of features is specified as 50. The model training is terminated when the training loss stabilizes (change less than 0.001).

During the model training, the input parameters consist of the previously mentioned through-casing transient electromagnetic logging curves and conventional well logging data. The output results comprise formation resistivity well log curves. The three wells are uniformly segmented into training, validation, and testing sets. Specifically, the data is segmented into three sections based on depth order, with the first 70%, middle 10%, and last 20% forming the training, validation, and testing sets in each section. It is noteworthy that the validation set is employed to evaluate the model after each epoch, and the model parameters from the epoch with the highest validation accuracy are selected as the final model parameters. The final dataset partition ratio is as follows: training set: validation set: testing set = 0.7:0.1:0.2, as illustrated in the figure below.

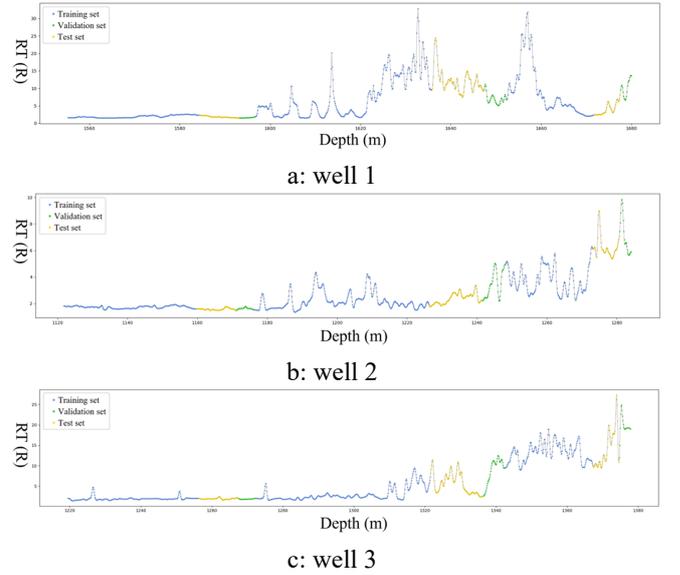

a: well 1

b: well 2

c: well 3

**Fig. 8.** Data partitioning results (using RT as an example)

*C. Experiment on Predicting Missing Well Logs*

This study addresses the problem of predicting formation resistivity using through-casing transient electromagnetic curves. Frequency-aware framework and temporal anti-noise block are proposed and combined to form the Frequency aware LSTM (FAL). The application scenario is the prediction of missing segments in the well log curve of formation resistivity.

To elucidate the working principles of the model and analyze the contributions of each component, ablation experiments were conducted. The purpose of ablation experiments is to selectively remove certain components or features from the model to evaluate their impact on model performance. By gradually eliminating factors from the model, researchers can delve into the model's robustness, stability, and performance, thereby gaining a more comprehensive understanding of the model's behavior. In addition to FAL, ablation experiments include FAF, which incorporates the FAF only, and TAL, which includes the temporal anti-noise LSTM block only. Additionally, Attention-LSTM with conventional attention mechanism and Res-LSTM with a residual structure are included as baseline for a complete ablation experiment. The prediction results of FAF, TAL, FAL, and baseline on this dataset are presented in Figure 9 and Table II.

TABLE II
RESULTS OF ABLATION STUDY

| Model | Training Loss | $R^2$ | MAE | RMSE | MSE | training time (s) | prediction time (s) |
|---|---|---|---|---|---|---|---|
| LSTM | 0.0253 | 0.69 | 1.90 | 2.62 | 6.86 | 1.7012 | 0.0435 |
| Attention-LSTM | 0.0135 | 0.74 | 1.64 | 2.35 | 5.54 | 2.9954 | 0.0495 |
| Res-LSTM | 0.0102 | 0.75 | 1.44 | 2.34 | 5.496 | 1.7698 | 0.0421 |
| FAF | 0.0037 | 0.80 | 1.21 | 2.07 | 4.29 | 15.2112 | 0.0644 |
| TAL | 0.0023 | 0.81 | 1.40 | 2.01 | 4.04 | 46.7555 | 0.0563 |
| **FAL** | **0.0027** | **0.91** | **0.95** | **1.40** | **1.96** | **31.4587** | **0.0536** |

Comparing to the baseline, it can be observed from the predictions of the third column subfigures in Figure 9 that FAF provides a more accurate fit to the actual values in the high-frequency curve section. However, its performance in the low-frequency domain shows little difference from that of conventional models, indicating that the FAF primarily focuses on the extraction and learning of high-frequency features. In contrast to the baseline, TAL enhances the accuracy across the board, with the most significant improvements seen in the predictions for low-frequency and low-amplitude segments (as depicted in the second column of Figure 9). This demonstrates that the temporal anti-noise block is capable of reducing the impact of noise in both high- and low-frequency bands, although its effectiveness in the high-frequency section is far less than that in the low-frequency section, due to the high-frequency disaster effect. Compared to the base line, FAL exhibits the best prediction performance in all subfigures of Figure 9. It not only maintains stability in the low-frequency band without being affected by noise interference but also demonstrates the same high-frequency extraction and learning capabilities as FAF in the high-frequency section, highlighting the model's resistance to interference and its ability to combat high-frequency disaster in the presence of strong noise and high redundant time-series data.

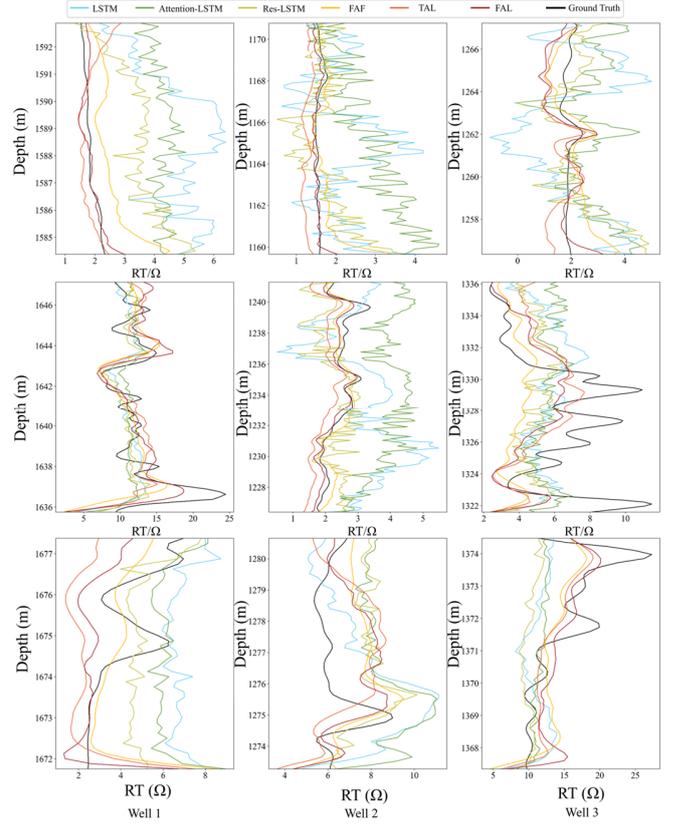

**Fig. 9.** Prediction results and comparison results

Table II presents the results of the ablation experiments (the bold part is our model), with each model group displaying Training Loss, training accuracy, training time, and prediction time. The training accuracy includes four metrics: $R^2$, MAE, RMSE, and MSE, with both overall $R^2$ and individual well $R^2$ values presented in the $R^2$ metric. For FAF, TAL, and FAL, there is a relative improvement of 0.1120, 0.1272, and 0.2212 in the overall $R^2$ compared to LSTM. Additionally, notable improvements are observed when compared to Attention-LSTM and Res-LSTM. These results provide evidence of the effectiveness of the FAF and temporal anti-noise block. To compare the proposed model's ability to address the high-frequency disaster, the prediction results are divided into high-frequency and low-frequency components, as shown in Figure 10 and Table III.

TABLE III
RESULTS OF MISSING WELL LOGS SEGMENTS PREDICTION IN DIFFERENT FREQUENCY DOMAIN

| Model | Frequency domain | $R^2$ | MAE | RMSE | MSE |
|---|---|---|---|---|---|
| LSTM | low | 0.5109 | 1.8553 | 2.5395 | 6.4490 |
|  | high | 0.4245 | 0.4197 | 1.0175 | 1.0352 |
| FAF | low  | 0.5518 | 1.7842 | 2.4013 | 6.0051 |
|     | high | 0.7248 | 0.3391 | 0.5329 | 0.5668 |
| TAL | low  | 0.8136 | 1.3787 | 1.9533 | 3.8153 |
|     | high | 0.4871 | 0.3229 | 1.02   | 1.0404 |
| **FAL** | **low**  | **0.8939** | **0.9696** | **1.464** | **2.1433** |
|     | **high** | **0.7937** | **0.3536** | **0.5223** | **0.5596** |

Figure 10 illustrates the performance of the proposed model and LSTM in predicting the high-frequency and low-frequency components of the results, while Table III provides specific error values for the models. Analyzing Figure 10 reveals that in the first-row subgraph of Figure 10(a), FAF demonstrates an advantage in predicting the high-frequency components, leading to an overall improvement of 0.2139 in R2 compared to LSTM in the high-frequency part. In the first-row subgraph of Figure 10(b), TAL shows improvement in countering noise interference in the low-frequency range, resulting in an overall improvement of 0.3027 in the low-frequency part. Overall, FAL exhibits the advantages of FAF in the high-frequency part (as evident in the first-row subgraph of Figure 10(a)) and TAL in the low-frequency part (as evident in the first-row subgraph of Figure 10(b)).

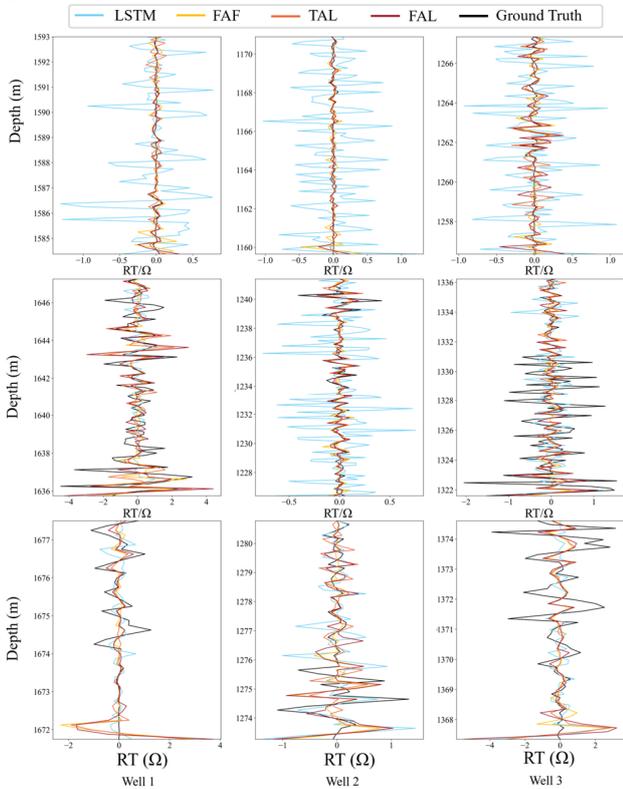

a: results in high frequency domain

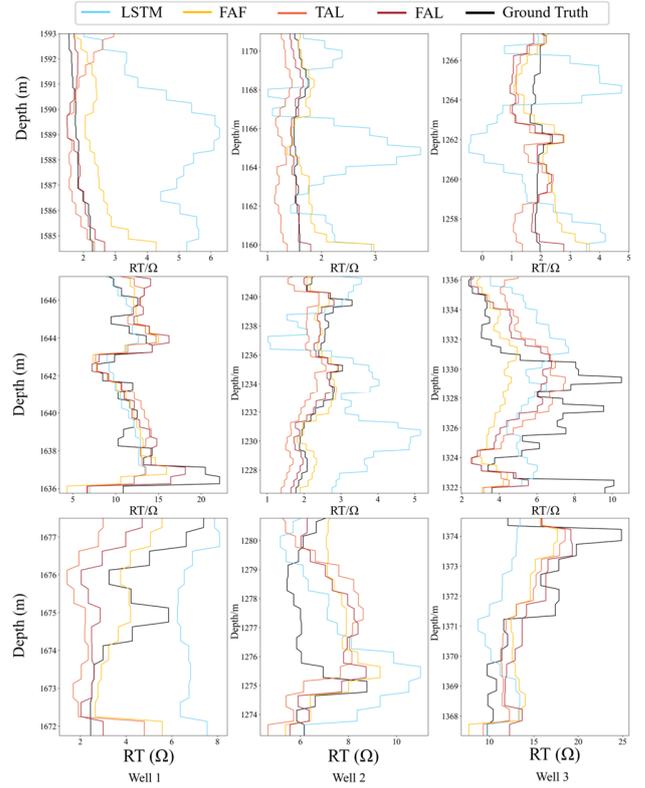

b: results in low frequency domain

**Fig. 10.** Result of missing well logs prediction in different frequency domain

### D. Robustness Analysis with Noisy Data

To compare the predictive capabilities of our proposed model with conventional models under complex underground well conditions, we simulated two common types of underground well noise scenarios: (1) Gaussian noise, simulating white noise and electronic noise inherent to logging instruments, such as mechanical noise, thermal noise, and flicker noise from logging equipment. (2) Impulse noise, simulating transient high-amplitude pulse noise generated during the logging process due to environmental factors, such as electromagnetic interference in the wellsite environment, including power grid frequency harmonics, and noise from radio transmitters. The formula for added Gaussian noise is expressed as follows.

$$p_i(z_i) = \frac{1}{\sigma\sqrt{2\pi}} e^{-\frac{(z_i - \mu)^2}{2\sigma^2}} \quad (9)$$

$$L_i' = L_i + \alpha p_i(z_i) \quad (10)$$

Where $p_i(z_i)$ represents Gaussian noise following a normal distribution at depth i. $\mu$、$\sigma$ denote the mean and variance of the well log curve to which noise is added, and $z_i$ is a uniformly random value between 0 and 1 at depth i. $\alpha$ is the scaling factor used to amplify the noise impact. $L_i'$ represents the well log curve value after adding noise at depth i, and $L_i$ represents the well log curve value before adding noise at depth i. The added impulse noise is expressed by the following equation.

$$n_i(z_i) = \begin{cases} \frac{1}{\sigma\sqrt{2\pi}} e^{-\frac{(z_i-\mu)^2}{2\sigma^2}} & z_i \leqslant \frac{1}{T} \\ 0 & z_i > \frac{1}{T} \end{cases} \quad (11)$$

$$L_i' = L_i + \alpha n_i(z_i) \quad (12)$$

Where $n_i(z_i)$ represents impulse noise at depth i. $\mu$ and $\sigma$ denote the mean and variance of the well log curve to which noise is added, and $z_i$ is a uniformly random value between 0 and 1 at depth i. $\alpha$ is the scaling factor used to amplify the noise impact. $L_i'$ represents the well log curve value after adding noise at depth i, and $L_i$ represents the well log curve value before adding noise at depth i, while $T$ represents the pulse period (set to 0.625m in this study).

We incorporated the addition of two types of noise into the well log curves of the input data. Our model was then compared with the baseline model to analyze their respective capabilities and sensitivities to noise interference. Four types of added noise were considered, namely: Gaussian noise (with $\alpha$ of 0.1), Gaussian noise (with $\alpha$ of 0.2), Impulse noise (with $\alpha$ of 0.1), and Impulse noise (with $\alpha$ of 0.2). The results are presented in Figures 11 and 12.

Comparing Figure 10 with Figure 11, it is observed that the prediction curve of LSTM undergoes the largest amplitude variations and resembles the trend of added noise, indicating that this neural network model is most affected by the two types of noise data during training. In contrast, the TAL and FAL models exhibit curves with amplitudes closer to the target curve, suggesting that the temporal anti-noise block can to some extent counteract Gaussian and impulse noise.

Table IV presents the performance of models in data with noise (the bold part is our model). Analyzing Table IV reveals that the $R^2$ of the LSTM model decreases the most under the influence of Gaussian noise and impulse noise, by 16.9% and 10.4%, respectively. The proposed models show the following reductions: FAF (Gaussian noise: 8.0%, impulse noise: 5.0%), TAL (Gaussian noise: 3.3%, impulse noise: 2.2%), and FAL (Gaussian noise: 2.2%, impulse noise: 1.6%). This comparative experiment demonstrates that the proposed models can effectively counteract both Gaussian and impulse noise.

TABLE IV
RESULTS OF MISSING WELL LOGS SEGMENTS PREDICTION WITH NOISE

| Model | Noise | Training Loss | $R^2$ | MAE | RMSE | MSE |
|---|---|---|---|---|---|---|
| LSTM | Gaussian noise ($\alpha$ of 0.1) | 0.0144 | 0.6453 | 2.149 | 2.8042 | 7.8635 |
|  | Gaussian noise ($\alpha$ of 0.2) | 0.0186 | 0.5729 | 2.3445 | 3.0767 | 9.4664 |
|  | Impulse noise ($\alpha$ of 0.1) | 0.0111 | 0.6513 | 2.1123 | 2.7801 | 7.729 |
|  | Impulse noise ($\alpha$ of 0.2) | 0.0082 | 0.6180 | 2.2495 | 2.91 | 8.4684 |
| FAF | Gaussian noise ($\alpha$ of 0.1) | 0.0043 | 0.7684 | 1.3313 | 2.2657 | 5.1336 |
|  | Gaussian noise ($\alpha$ of 0.2) | 0.0043 | 0.7418 | 1.6476 | 2.3925 | 5.7241 |
|  | Impulse noise ($\alpha$ of 0.1) | 0.0043 | 0.7905 | 1.4514 | 2.1551 | 4.6443 |
|  | Impulse noise ($\alpha$ of 0.2) | 0.0049 | 0.7663 | 1.5938 | 2.276 | 5.18 |
| TAL | Gaussian noise ($\alpha$ of 0.1) | 0.0047 | 0.7972 | 1.3494 | 2.1204 | 4.4963 |
|  | Gaussian noise ($\alpha$ of 0.2) | 0.0031 | 0.7904 | 1.4514 | 2.1556 | 4.6466 |
|  | Impulse noise ($\alpha$ of 0.1) | 0.0022 | 0.8051 | 1.4456 | 2.0785 | 4.32 |
|  | Impulse noise ($\alpha$ of 0.2) | 0.0018 | 0.7990 | 1.3477 | 2.1107 | 4.4552 |
| **FAL** | **Gaussian noise ($\alpha$ of 0.1)** | **0.0032** | **0.9003** | **1.0198** | **1.4863** | **2.2091** |
|  | **Gaussian noise ($\alpha$ of 0.2)** | **0.0031** | **0.8908** | **1.0501** | **1.5558** | **2.4205** |
|  | **Impulse noise ($\alpha$ of 0.1)** | **0.0029** | **0.9009** | **1.0224** | **1.4825** | **2.1977** |
|  | **Impulse noise ($\alpha$ of 0.2)** | **0.0030** | **0.8965** | **1.0352** | **1.5149** | **2.2949** |

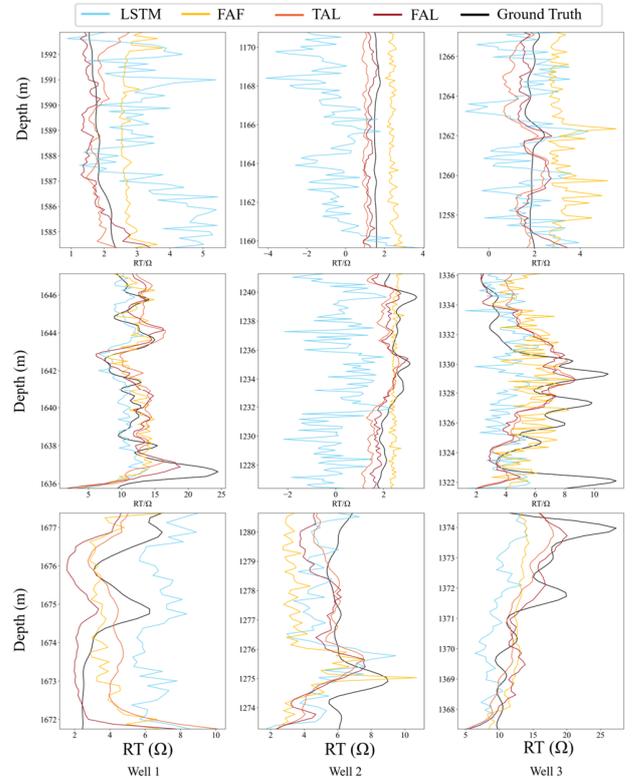

a: results of data with Gaussian noise of 0.1 in $\alpha$

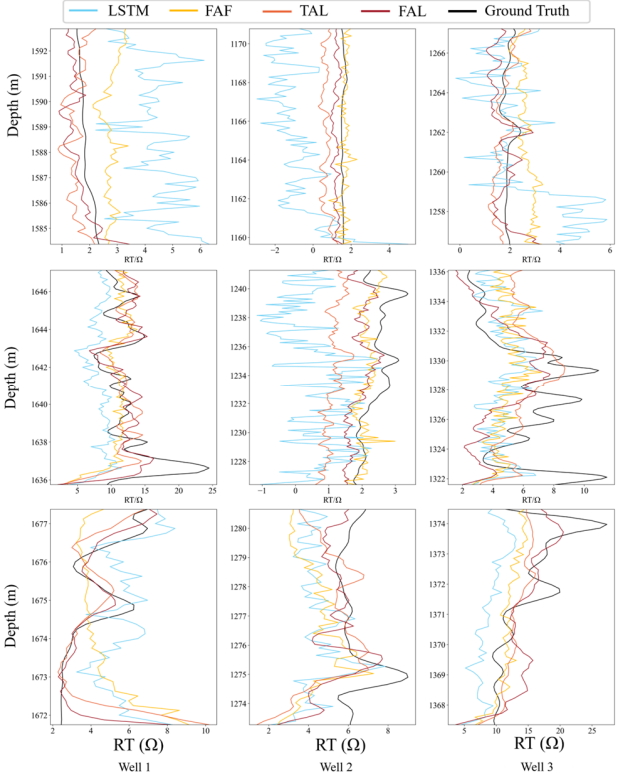

b: results of data with Gaussian noise of 0.2 in $\alpha$

**Fig. 11.** Result of missing well logs prediction with Gaussian noise

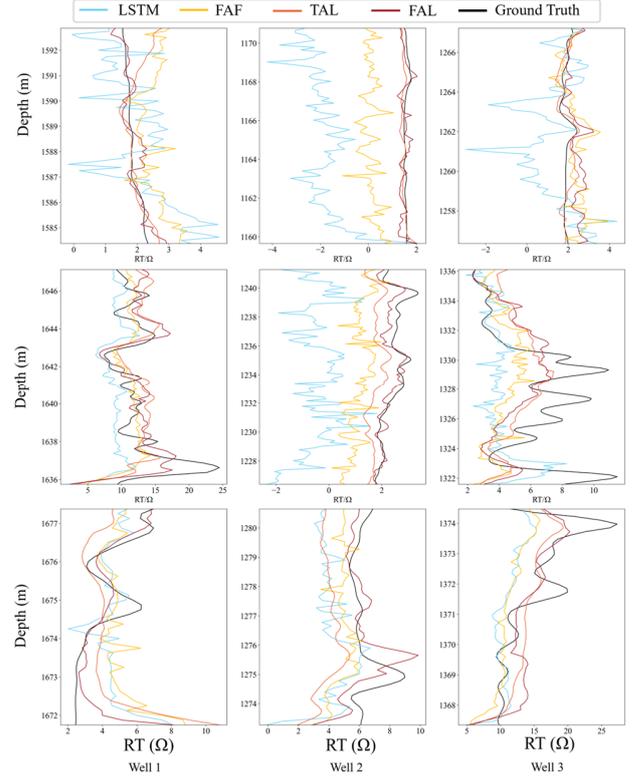

b: results of data with Impulse noise of 0.2 in $\alpha$

**Fig. 12.** Result of missing well logs prediction with Impulse noise

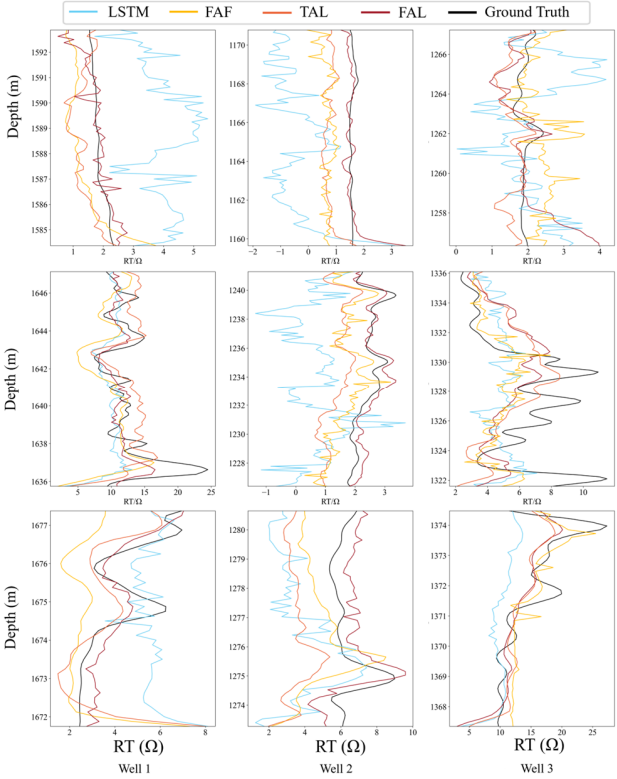

a: results of data with Impulse noise of 0.1 in $\alpha$

## IV. CONCLUSIONS

In this study, to address the challenge of predicting formation resistivity based on the over cased transient electromagnetic well log curves, FAF and temporal anti-noise block are proposed. The aims of these two methods are to address two problems:(1) the high-frequency disaster (2) temporal data noise, and high redundancy issues. In terms of model validation, two experiments are conducted: ablation experiment in missing well logs prediction and robustness analysis with noisy data.

The results indicate that in the ablation experiment for well logs missing points prediction, FAL outperforms LSTM, Attention-LSTM, and Res-LSTM by improving $R^2$ values by 24.3%, 17.7%, and 17.4%, respectively. FAL achieves the highest $R^2$ of 0.9113 among all models, demonstrating the model's enhanced prediction accuracy. Furthermore, the analysis of the prediction results for the high-frequency and low-frequency components individually validates the effectiveness of the model in combating high-frequency disaster through its multiscale structure. In the Performance of robustness analysis with noisy data, FAL exhibits only a 2.2% and 1.6% reduction in $R^2$ under the influence of Gaussian and impulse noise, respectively. This reduction is approximately 1/8 of the reduction observed in the baseline, confirming the noise resistance of the proposed model. Experiments have found that independently learning the high and low frequency components in time-series data can avoid high-frequency disaster, thereby improving the learning ability and prediction accuracy of neural networks in high-frequency components; In addition, the combination of multi head attention mechanism and soft threshold attention mechanism can effectively reduce the influence of noise and redundancy in temporal data, thereby reducing the influence of random noise through soft threshold mechanism.